%% file: main.tex
\documentclass{article}

\PassOptionsToPackage{}{natbib}
\usepackage[final]{neurips_2022}

\usepackage{subcaption}
\usepackage{amsmath,xspace}
\usepackage{multirow}
\usepackage{amssymb}
\usepackage{graphicx,paralist}
\usepackage[export]{adjustbox}
\usepackage[utf8]{inputenc} % allow utf-8 input
\usepackage[T1]{fontenc}    % use 8-bit T1 fonts
\usepackage{hyperref}       % hyperlinks
\usepackage{url}            % simple URL typesetting
\usepackage{booktabs}       % professional-quality tables
\usepackage{amsfonts}       % blackboard math symbols
\usepackage{nicefrac}       % compact symbols for 1/2, etc.
\usepackage{microtype}      % microtypography
\usepackage{xcolor}         % colors
\usepackage{colortbl}
\usepackage{subcaption}
\usepackage{ulem}

\usepackage{tabularx}

\definecolor{MyDarkBlue}{rgb}{0,0.08,1}
\definecolor{MyDarkGreen}{rgb}{0.02,0.6,0.02}
\definecolor{MyDarkRed}{rgb}{0.8,0.02,0.02}
\definecolor{MyDarkOrange}{rgb}{0.40,0.2,0.02}
\definecolor{MyPurple}{RGB}{111,0,255}
\definecolor{MyRed}{rgb}{1.0,0.0,0.0}
\definecolor{MyGold}{rgb}{0.75,0.6,0.12}
\definecolor{MyDarkgray}{rgb}{0.66, 0.66, 0.66}

\newcommand{\cready}[1]{#1}

\newcommand{\rb}[1]{\textcolor{MyRed}{\textbf{#1}}}
\newcommand{\bb}[1]{\textcolor{MyDarkBlue}{\textbf{#1}}}
\newcommand{\gb}[1]{\textcolor{MyDarkGreen}{\textbf{#1}}}
\newcommand{\pb}[1]{\textcolor{MyPurple}{\textbf{#1}}}

\title{Controllable Text Generation with Neurally-Decomposed Oracle}
\author{
  Tao Meng\thanks{equal contribution}\\
  University of California, Los Angeles\\
  \texttt{tmeng@cs.ucla.edu} \\
  \And
  Sidi Lu\footnotemark[1]\\
  University of California, Los Angeles\\
  \texttt{sidilu@cs.ucla.edu} \\
  \AND
  Nanyun Peng\\
  University of California, Los Angeles\\
  \texttt{violetpeng@cs.ucla.edu} \\
  \And
  Kai-Wei Chang\\
  University of California, Los Angeles\\
  \texttt{kwchang@cs.ucla.edu} \\
  }

\begin{document}

\def\prefix#1{\mathbf{y}_{<#1}}
\def\prefixeq#1{\mathbf{y}_{\leq#1}}
\newcommand{\bxy}{\mathbf{x,y}}
\newcommand{\byx}{\mathbf{y|x}}

\maketitle

\begin{abstract}

%We propose controllable text generation with NeurAlly-Decomposed Oracle (NADO), a general and efficient framework for controlling auto-regressive generation models by decomposing sequence-level oracle. 
We propose a general and efficient framework to control auto-regressive generation models with NeurAlly-Decomposed Oracle (NADO).
Given a pre-trained base language model and a sequence-level boolean oracle function, we propose to
decompose the oracle function into token-level guidance to steer the base model in text generation. 
%token-level to generate sequence satisfying the oracle function. 
Specifically, the token-level guidance 
%re-weighting token distribution 
is approximated by a neural model trained with examples sampled from the base model, demanding no additional auxiliary labeled data. 
Based on posterior regularization, we present the closed-form optimal solution to incorporate the token-level guidance into the base model for controllable generation. 
%We formulate the control as an optimization problem taking oracle as constraints, and present the optimal solution of incorporating the decomposed token-level guidance. 
%We further discuss how the neural approximation affects the quality of the solution.
We further provide a theoretical analysis of how the approximation quality of NADO affects the controllable generation results.
Experiments conducted on two tasks: (1) text generation with lexical constraints and (2) machine translation with formality control demonstrate that our framework efficiently guides the base model towards the given control factors while maintaining high generation quality.
%, without refacting or fine-tuning the base model. 

\end{abstract}
% \sidi{MEnD: Mask-Enforcing Decoding for Controllable Text Generation with Large Models}

% \sidi{Tao and I currently prefer this one ↑. MEnD as in "we are \emph{mend}ing the distribution of generated samples from unsatisfying constraints."}

% \sidi{CEnD: Constraint-Enforcing Decoding for Controllable Text Generation with Large Models}
% \sidi{CEnT: Constraint-Enforcing Training bla bla bla}

% \sidi{MEnD: Mask-Enacting Decoding for Controllable Text Generation with Large Models}

% \sidi{MInD: Mask-Instructed Decoding for Controllable Text Generation with Large Models}

% \sidi{MEnCius: Mask-Enforcing ControllIng for UnSatisfied Constraints in Text Generation with Large Pretrained Models}

% \tao{WE NEED A NAME! PR + *Distribution + *Decoding + }
% \tao{oracle name}
\input{introduction}

\input{related}

\input{methodology}

\input{experiment}

\section{Conclusion}
  %  Version 1: In this work we propose a general and efficient framework for controllable generation. We generally formulate the controlling code as a sequence-level boolean oracle function, which is compatible for various application scenarios. By decomposing the sequence-level oracle into token-level guidance, NADO efficiently guides the base model by two parallel forward passes in the generation. The close-form optimal solution is approximated by NADO trained on data sampled from the base model. Therefore, NADO does not require auxiliary data or other external oracles for training the auxiliary model, avoiding the distributional discrepancy issue between base model and the NADO. We further provide related analysis about relation between errors in NADO and in sequence-level distribution. Experiment results support our claim and significantly improve the generation quality compared to existing work. We also provide solution for controlling with soft constraints, which could potentially applied in boarder scenarios. For example, reducing gender bias in generation by gender-related oracle and soft controlling.

We purpose a general and efficient framework for controllable generation. We leverage an auxiliary neural model, NADO, to approximate the decomposed oracle guidance, and incorporate it with a fixed base model. %Although it is not a novel idea to train a auxiliary model to guide the large base model, in this work we would like to highlight the effect of sampling. 
By training with sampled data from the base model, NADO aligns better with the base model, and our framework is more flexible dealing with various application scenarios provided by different sampling methods. As NADO is a general framework, in the future, we plan to apply it in boarder application scenarios. For example, reducing societal bias~\citep{sheng2019woman} (e.g., gender or racial bias) in generation by providing corresponding oracle.
%In the future, we plan to 
%further analyze the properties of NADO, for example, what kind of oracles can be easily decomposed and approximated. We will also apply soft controlling in boarder scenarios. For example, reducing gender bias in generation by gender-related oracle.

\section*{Acknowledgement}

We thank anonymous reviewers for their comments and suggestions. 
We also thank members at UCLANLP and UCLAPLUS labs for their feedback. The project is supported in part by 
CISCO, Amazon, Sloan foundation, Google, and DARPA MCS program under Cooperative Agreement N66001-19-2-4032. In addition, Tao is supported as an Amazon Fellow. 

\section*{Limitation}
    In this work we assume that a base model with decent quality (e.g., large pretrained language models) and a good oracle for controlling attributes are available. However, in some applications, the quality of the base model may be low and the oracle may only capture superficial shortcut between constraints and labels. How to control a generation model under these situations is an interesting future work direction. 

    Similar to other language generation approaches, we note that there is a risk that malicious users may use NADO to generate improper or toxic texts. Also, the generated text may contain societal biases inherited from data. However, on the other hand, NADO provides a powerful weapon against toxicity as developers can design constraints to detoxify the generated text. We refer readers to the discussion in \cite{sheng2019woman,sheng2021societal,zellers2019defending,bender2021dangers,radford2019gpt2, brown2020gpt3,dev2021harms,dhamala2021bold}.

\clearpage
\bibliography{ref}
\bibliographystyle{acl_natbib}

\input{checklist}

\appendix
\input{appendix}

\end{document}

%% file: introduction.tex
\section{Introduction}
    Auto-regressive language models have been widely used for text generation. 
    With the recent development of large-scale pre-trained language models~\citep{radford2019gpt2,brown2020gpt3,raffel2020t5, lewis2020bart},
    %Specifically, the sequence-to-sequence learning framework has been applied to a wide variety of 
    they have achieved state-of-the-art performances in applications such as machine translation \citep{bahdanau2015nmt, luong2015effective}, image captioning \citep{anderson2018bottom, you2016image} and open domain text generation~\citep{zhang2014chinese,yao2019plan,vinyals2015neuraldialogue,shang2015neural,lu2018neural}. 
    %In models with such formulation, control over the generated contents is achieved by conditioning the tractable estimation of output data probability function on the given input. Such models achieve controlled generation through learning from existing samples, which best suit the cases where parallel data is available. \sidi{add ref.}
    %\tao{Motivation / Goal: Control large generation models. Fine-tune is slow.}
    %However, in application scenarios, we probably don't have sufficient parallel data to well-specify the control we want to enforce.
    %In many applications, we expect 
    However, many applications such as open-domain creative generation~\citep{yao2019plan,goldfarb2020content,tian2022sonnet,han2022go,hong2022Character,spangher2022sequentially} require to control model output with specific sequence-level attributes. The attributes can be specified by a set of rules\footnote{For example, lexical constraints require certain words to appear in the generated text~\citep{hokamp2017lexical, lin2019comgen}} or by an abstract concept (\emph{e.g.}, the generated text follows a particular writing style).
    How to control auto-regressive language models to satisfy these attributes is an open challenge.
    
    %Hence, we seek for a method to control the model outputs in the decoding stage. However, it could be challenging to control an auto-regressive model since it predicts the distribution over all possible tokens and generate one of them per step. What we require is the token-level guidance to gear the token distribution rather than the sequence-level supervision.
    
    %\tao{What we do}
      
    % \begin{figure}[t]
    %     \centering  
    %     \includegraphics[width=0.8\linewidth]{PipelineIllustration.pdf}
    %     \caption{An illustration of NADO. We use an auxiliary model to approximate the close-form solution trained by examples sampled from base model. We incorporate auxiliary model with base model to form an oracle-injected model to generate texts.}
    %     \label{fig:intro}
    % \end{figure}
    
    \begin{figure}[t]
        \centering  
        \begin{subfigure}{0.46\linewidth}
            \includegraphics[width=\linewidth]{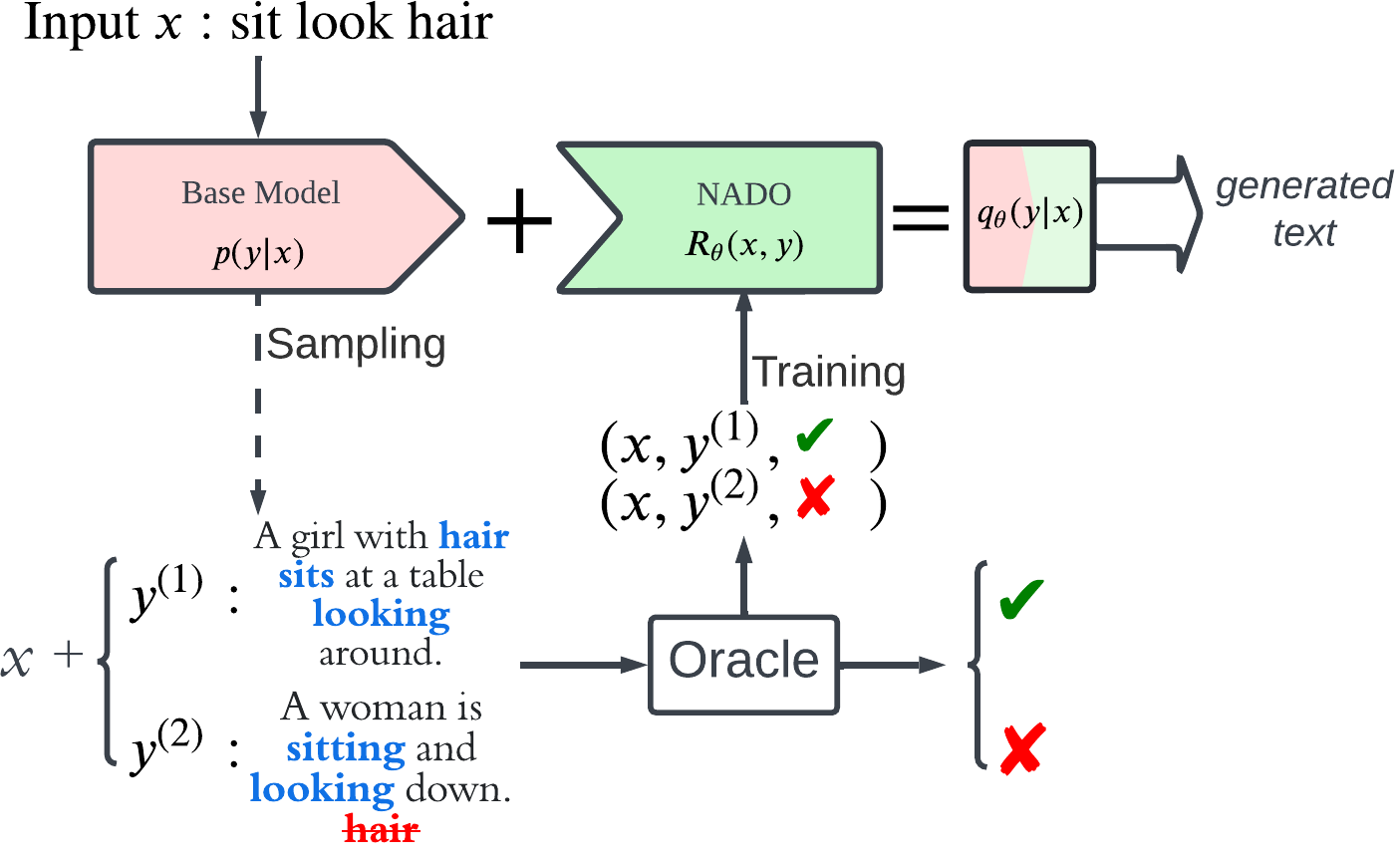}
            %\caption{An illustration of NADO. Here we take lexical constrained generation as an example, and the oracle is checking whether all keywords in the input $\mathbf{x}$ are incorporated in generated text $\mathbf{y}$. We sample examples from base model $p$ and feed them into the oracle. With proper training with the sampled examples, the oracle is decomposed into token-level and approximated by auxiliary model $R_\theta$. The base model together with the auxiliary model forms an oracle-injected auto-regressive model.}
            \caption{Take lexically constrained generation as an example, where the oracle checks whether all keywords in the input $\mathbf{x}$ are incorporated in generated text $\mathbf{y}$. With proper training using samples from the base model $p$ (dashed arrow) labeled by the oracle, we decompose the oracle into token-level guidance and parameterize it by an auxiliary model $R_\theta$ (NADO). We use $R_\theta$ to provide guidance when generating text with the base model (see details in Fig. 1(b)).}
            \label{fig:intro_fashion}
        \end{subfigure}
        \hspace{0.2in}
        \begin{subfigure}{0.46\linewidth}
            \includegraphics[width=\linewidth]{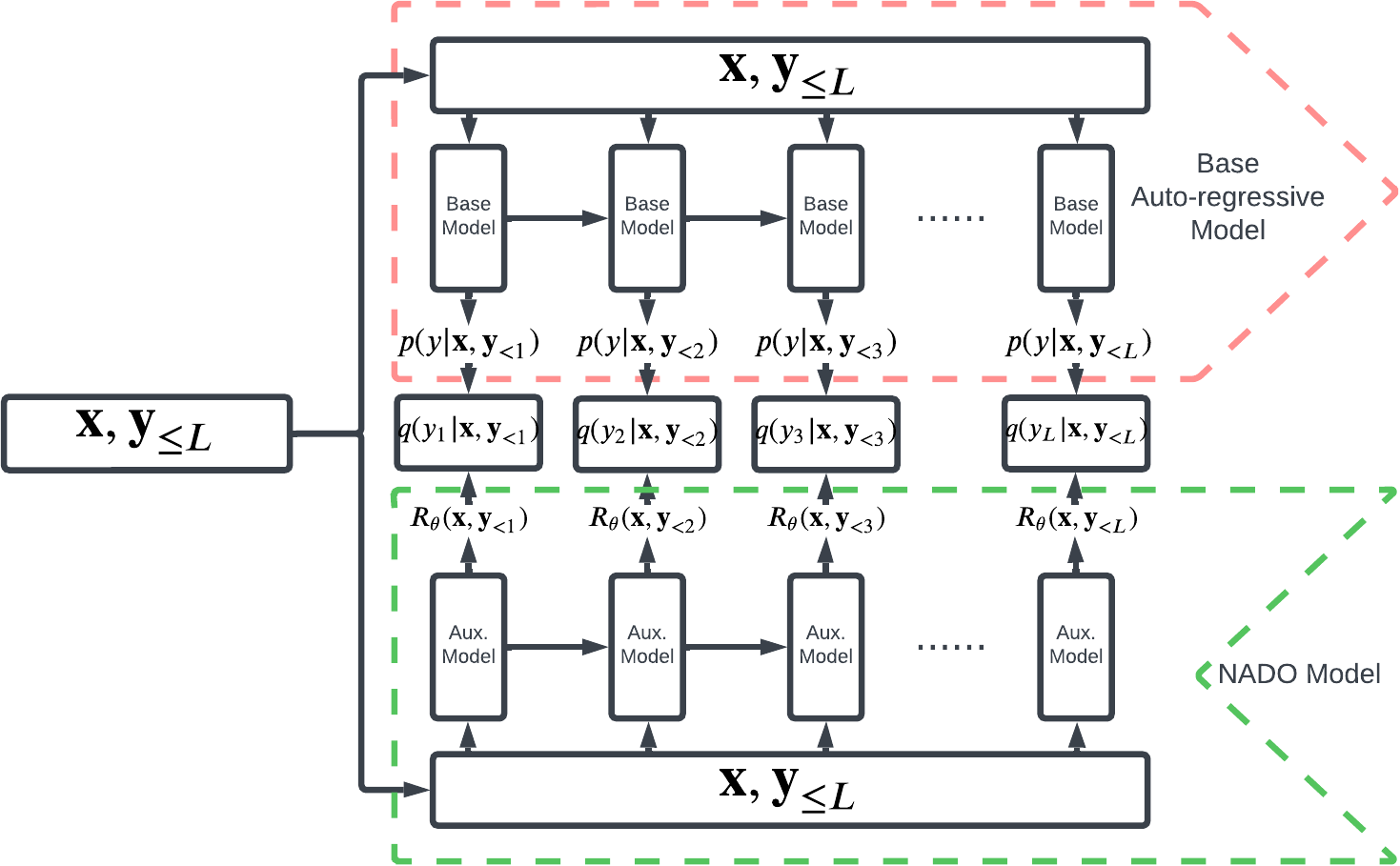}
            \caption{Illustration of the controlled generation process. Both the base model and the auxiliary model (NADO) take input $\mathbf{x}$ and the generated sequence (prefix) $\prefix L$ as input. The base model, in each step, outputs a token distribution $p(y_i|\mathbf{x},\prefix i)$. Guided by NADO $R_\theta$, we obtain the distribution $q$ (See Sec. \ref{sec:rcp}), based on which we generate the output token. 
            }
            \label{fig:rc}
        \end{subfigure}
        \caption{Illustration of pipeline incorporating NADO (left) and model architecture (right).}
    \end{figure}
    
    In this paper, we propose a general and flexible framework for controllable text generation. Given a base pre-trained language model and a sequence-level oracle function indicating whether an attribute is satisfied, our goal is to guide the text generation to satisfy certain attributes using the oracle. 
    To this end, we propose to decompose the sequence-level oracle into token-level guidance,  
    %We then define a feasible distribution set according to NADO, in which we seek for the closest distribution from the base auto-regressive model.
    such that when generating the $i-$th token in the output sequence given the prefix, instead of sampling from the base model, we modify the probability distribution of the output token based on the token-level guidance. 
    Specifically, we formulate the control as an optimization problem based on posterior regularization~\citep{ganchev2010posterior} and solve the closed-form optimal solution to incorporate the token-level guidance for text generation. The decomposition is approximated by an auxiliary neural network model, called NeurAlly-Decomposed Oracle (NADO), which is trained on data sampled from the base model and supervised by the sequence-level oracle (see the illustration Fig. \ref{fig:intro_fashion}). 
    We further provide theoretical analysis on how NADO's approximation quality affects the controllable generation results.
    %We then utilize an auxiliary model to approximate the solution. As shown in Fig. \ref, the auxiliary model is trained with data sampled from the base model. In decoding, we are able to efficiently incorporate the auxiliary model to the base model based on the close form token-level guidance solution. 
    Note that in the entire process, we treat the base model and the sequence-level oracle as black-box functions, without the need for any refactoring or fine-tuning.

    %Hence, in this paper, we propose a general and flexible framework for controllable generation. Specifically, we decompose the sequence-level attributes into token-level, called 
    %\tao{Previous work (neural logic, FUDGE, PPLM, GeDI, please add), we are better.}
    
    A few existing controllable generation works (\emph{e.g.},  \citet{lu2021neurologic,lu2022neurologic}) design search algorithms for generating texts with \textit{lexical constraints}. However, their approaches cannot generally be applied to constraints such as style.
    %specific trainable decoding algorithm and proved to be effective in lexical constrained generation tasks but lack of 
    %generalizability. 
    Another line of work such as PPLM \citep{dathathri2020pplm}, GeDI \citep{krause2021gedi}, and FUDGE \citep{yang2021fudge} also aim to guide the base model with an auxiliary model. However, they either shift the base model distribution in a post-hoc manner without theoretical guarantee, %an implicit but trainable strategy incorporating the attribute, 
    or/and require external labeled data to train the auxiliary model. \citet{khalifa2021distributional, korbak2022controlling} propose a generation with distributional control approach. Our control objective derived through posterior regularization resembles their energy-based model representation. However, they approximate the energy-based model using a KL-adaptive distributional policy, while we propose to decompose the sequence-level oracles into token-level approximated by NADO. With the decomposition, base models receive explicit controlling signal in generating every token from the oracle.
    %formulate the control objectives as an energy-based model and approximate it by a KL-adaptive distributional policy gradient method. 
%    optimal sentence-level distribution, and leverage reinforcement learning to approximate it.
    %Instead, we provide close-form optimal solution for incorporating the attributes, and exclusively require the sequence-level oracle, which avoids the distributional discrepancy issue between the base auto-regressive model and auxiliary guidance.
    %Instead, we derive a token-level closed-form decomposition solution for the optimal way to incorporate the oracle. %without requiring external labeled data or token-level guidance. 
    Furthermore, since NADO is trained on the data sampled from the base models, it aligns better with the base model's distribution and thus can achieve better control.
    
    We conduct experiments on lexically constrained generation (LCG) tasks and a machine translation (MT) formality change task. In LCG tasks, the oracle is a rule-based keyword checker. We achieve almost perfect keyword incorporation with significantly boosted BLEU scores compared to previous approaches that design specific decoding algorithms~\citep{lu2021neurologic}. 
    In the formality-controlled MT task, we are provided with a formality oracle predicting whether a sentence is formal or not, and the goal is to guide the model to generate formal translations. Compared with recent work~\citep{yang2021fudge}, we improve the BLEU score by 3 points as well as improve the formality rate, demonstrating NADO's superior ability to incorporate external oracle supervision. 
    %In both experiments, we are able to decompose the sequence-level oracle supervision into token-level guidance. 
    Both experiments demonstrate the effectiveness of our framework in dealing with various types of control while maintaining high-quality generation results.\footnote{Our code can be found at \url{https://github.com/MtSomeThree/constrDecoding}.}

%% file: related.tex
\section{Related Work}
    %\tao{Should we cover related work in RL?} \sidi{I think we'd better not as it may diverge the focus from our current story...}
    
     %\tao{Three types of controllable generation. We are in 3, (1,2) have disadvantage, we don't have disadvantage in 3.}
     \noindent \textbf {Controllable Text Generation with Auto-regressive Models.} %Auto-regressive models have long been considered the standard paradigm text generation. 
     Most previous work on controllable text generation are based on the auto-regressive framework. \citet{zhang2022survey} summarize %existing work 
     these methods into three categories: \emph{fine-tuning}, \emph{refactor/retraining} and \emph{post-processing}. The first two categories, e.g., fine-tuning with control code~\citep{peng2018towards,keskar2019ctrl} or prompt-based methods \citep{sheng2020towards,shin2020autoprompt,letster2021power,li2021prefix}, are usually weaker in controllability and inefficient in training considering the size of language models are dramatically increasing nowadays. Generally, the post-processing methods are considered expensive in inference and low quality in generated texts. However, our framework, as a kind of post-processing method, is able to achieve high generation quality demonstrated in the experiments and maintains efficient inference.
     
     %    Usually, we can easily access a sequence-level boolean oracle for those attributes. In auto-regressive models, controlling over the generated contents can be naturally achieved by fine-tuning or retraining the models on examples with specific attributes \citep{keskar2019ctrl}, including prompt tuning \citep{shin2020autoprompt,letster2021power,li2021prefix}. However, the sizes of pre-trained auto-regressive models, e.g., GPT-2\citep{radford2019gpt2} and GPT-3\citep{brown2020gpt3}, are dramatically increasing nowadays, and this type of controlling methods are considered inefficient in terms of both training time and model space.
     
     %\tao{Two class of post-processing, decoding algorithm is not fundamental, and can be used as complementary.}
     \noindent \textbf{Controllable Text Generation via Post-processing.} There are two major lines in post-processing: (1) modifying the decoding algorithm and (2) guiding generation with an auxiliary model. For some token-level controlled generation tasks like lexically constrained generation, we can inject the constraints into the decoding algorithm (e.g., constrained beam search \citep{anderson2017guided,post2018fast} and NeuroLogic decoding \citep{lu2021neurologic, lu2022neurologic}). Though shown effectiveness in lexically constrained generation,
     these algorithmic methods fail to fundamentally touch the token distribution, and are hard to handle other abstract attributes.
     %these algorithmic methods are hard to handle other abstract attributes.
     
     %\tao{Summarize similar work. Key difference: 1. we don't need auxiliary data or token-level oracle, instead, we use a sentence-level oracle. 2. We train the auxiliary model by sampled data rather than other labeled data ==> remove the distributional discrepancy.}
     %\paragraph{Differences from Existing Methods in Auxiliary Model Guided Decoding} 
     In the second line, PPLM \citep{dathathri2020pplm} proposes an auxiliary discriminator for the expected attribute to guide the model; GeDi \citep{krause2021gedi} and DEXPERTS \citep{liu2021dexpert} apply contrastive learning and train an auxiliary language model to reweight the token distribution in each step; Plug-and-Blend \citep{lin2021plug} further extends the GeDi framework by adding a planner architecture. FUDGE \citep{yang2021fudge} leverages external token-level oracle to train a discriminator for guiding the base model. 
     These methods either require external token-level oracle guidance or auxiliary labeled datasets to train the auxiliary models. However, the distribution of the data used to train the auxiliary model is different from what the based model is trained on. This distributional discrepancy causes the drop of generating quality as we will show in the experiments. For example, given a controlling attribute $a$, Fudge generates next token $y_i$ based on Bayesian rule $P(y_i|\prefix i,a)\propto P(a|\mathbf{y}_{\leq i})P(y_i|\prefix i)$. However, their $P(a|\mathbf{y}_{\leq i})$ and $P(y_i|\prefix i)$ are not estimated based on the same distribution. In contrast, NADO is trained with data sampled from the base model. Therefore, it learns to incorporate with the base model, which avoids the distributional discrepancy. We also provide a principle, theoretical framework to discuss the optimal solution of incorporating the sequence-level oracle.

%% file: methodology.tex
\section{Methodology}
    %In this section we describe our method and corresponding formulation. 
    We approach the sequence-level controllable text generation problem by decomposing the sentence-level oracle into token-level guidance. We formulate this as an optimization problem. Since the token-level guidance is intractable, we propose to train an auxiliary model, called NeurAlly-Decomposed Oracle (NADO), to approximate it. During the inference time, NADO guides the base model to generate sequences that satisfy the oracle constraints.
    %we train an auxiliary model incorporating the given oracle boolean function by data sampled from the given base model, predicting how likely the base model will generate a sequence satisfies the boolean function with given condition. This auxiliary model is able to guide the base model to re-weight the token distribution in each step. 
    
    In the rest of this section, we discuss 1) the formulation to decompose the sequence-level oracle function into token-level guidance; 2) the formulation to incorporate the token-level guidance into the base model to achieve control; 3) the approximation of the token-level guidance using NADO; 4) a theoretical analysis of the impact of NADO approximation to the controllable generation results; and 5) the training of NADO.
    %We organize this section by answering the following four questions: (1) How to decompose the sequence-level oracle function into token-level guidance? (2) How to define NADO and incorporate it with the base model?  (3) How to train NADO? (4) How the NADO quality affects the generation results? 
    %We first introduce our notations and formalize our goal, and then describe how we incorporate the auxiliary model to the base model, with related theoretical analysis. After that we introduce how we train this auxiliary model, with various of sampling strategies for different application scenarios.
    
    \subsection{Setup: Notations and Problem Formulation}
    \label{sec:goal}
        We use $\mathbf{x}\in \mathcal{X}$ to denote the input and $\mathbf{y}\in \mathcal{Y}$ to denote the generated sequence. $y_i$ is the $i-$th token in $\mathbf{y}$ and $\prefix i$ is the sequence prefix from the beginning to the $(i-1)-$th token. We denote the base auto-regressive generation model as
        $p(y_i|\mathbf{x}, \prefix i),$ hence the sequence-level distribution is given by $p(\byx)=\prod_i p(y_i|\mathbf{x}, \prefix i).$ A sequence-level oracle is defined as a boolean function $C:\mathcal{X} \times \mathcal{Y}\rightarrow \{0,1\}.$ We formalize the optimization objective based on posterior regularization \citep{ganchev2010posterior}. Basically, we explore a token-level distribution $q^*(y_i|\mathbf{x}, \prefix i)$ and its corresponding sequence-level distribution $q^*(\byx)$, satisfying
    
    \begin{enumerate}%[label=(\roman*)]
        \item $q^*(\byx)=\prod_i q^*(y_i|\mathbf{x}, \prefix i), $ \emph{i.e.}, $q^*$ can be treated as an auto-regressive model.
        \item $q^*(\byx)=0$ if $C(\bxy)=0$, \emph{i.e.}, $q^*$ only generates sequences satisfying the oracle $C.$
        \item Given an input $\mathbf{x}$, $KL(p(\byx) \|  q^*(\byx))$ is minimized, \emph{i.e.}, $q^*$ should be as similar to the base model as possible. 
    \end{enumerate}
    
    \citet{khalifa2021distributional, korbak2022controlling} derive a similar optimization formulation as property 2, 3, 
    to represent constraints through energy-based models and approximate it with distributional policy gradient.
    In this work, we propose to decompose oracle to token-level guidance to steer the generation. 
    We discuss our approach in the following.
    %and control the model with reinforcement learning. Instead, in this work we focus on decomposing the oracle to token-level guidance
    %, and control the base model in an offline manner.
    
    \subsection{Token-level Guidance and Closed-Form Solution For $q^*$}
    \label{sec:rcp}
        Before we compute the solution for $q^*$,  given the base model $p$ and oracle $C$, we first define the token-level guidance as a success rate prediction function $R^C_p(\mathbf{x})$, which defines the probability of the sequence generated by $p$ satisfies the oracle $C$ given the input $\mathbf{x}$. We similarly define $R^C_p(\mathbf{x},\prefixeq{i})$ as the probability of success given input $\mathbf{x}$ and prefix $\prefix{i}$. By definition, we have
        
        \begin{equation}
        \label{eq:rc}
            \begin{aligned}
                R^C_p(\mathbf{x})&=\Pr\nolimits_{\mathbf{y}\sim p(\byx)}\left[C(\bxy)=1\right]=\sum\nolimits_{\mathbf{y}\in\mathcal{Y}}p(\byx)C(\bxy) \\
                R^C_p(\mathbf{x},\prefixeq{i})&=\Pr\nolimits_{\mathbf{y}\sim p(\byx)}\left[C(\bxy)=1|\prefix{i}\right]=\sum\nolimits_{\mathbf{y}\in\mathcal{Y}}p(\byx, \prefix{i})C(\bxy).
            \end{aligned}
        \end{equation}

        With the function $R^C_p$, we now derive the closed-form solution of $q^*$ considering conditions 2 and 3 defined in Sec. \ref{sec:goal}. 
        Given input $\mathbf{x}$, we define the feasible sequence-level distribution set $Q$ as
        \begin{equation}
        \label{eq:Q}
            Q:=\{q|\sum\nolimits_{\mathbf{y}:\ C(\bxy)=0} q(\byx)=0\},
        \end{equation}
        then the sequence-level closed-form solution for $q^*$ is given by
        \begin{equation}
        %\small
        \label{eq:seqlvl}
            q^*(\mathbf{y|x)}=\arg\min_{q\in Q} KL(p(\byx)\|q(\byx))=\frac{p(\mathbf{y|x)}C(\bxy)}{R^C_p(\mathbf{x})}.
        \end{equation}
        
        Considering condition 1 in Sec. \ref{sec:goal} to make $q^*$ tractable, we decompose $q^*(\byx)$ into token-level. The closed-form solution is given by
        
        \begin{equation}
        %\small
        \label{eq:toklvl}
            q^*(y_i|\mathbf{x},\prefix{i})=\frac{R^C_p(\mathbf{x},\prefixeq{i})}{R^C_p(\mathbf{x},\prefixeq{i-1})}p(y_i|\mathbf{x},\prefix{i}).
        \end{equation}
    
        The decomposition is unique. The proof and detailed derivation can be found in the appendix. 
        
        \noindent \textbf{Control with Soft Constraints.} In Eq. \eqref{eq:Q} we define the feasible distribution set as distribution that the possibility of a sequence violate the oracle function is $0$. However, in some applications, we expect to control the generation with soft constraints. For example, we want the model to generate sentence about sports with probability $r=0.8$. Our framework also supports controlling the generation with soft constraints. To achieve this, with a pre-defined ratio $r\in [0,1]$, we alternatively define a general feasible set $Q$ as
        %We are also capable to relax the constraint that with some pre-defined ratio $r\in [0,1]$, we relax feasible set $Q$ as 
        \begin{equation*}
        %\small
            Q:=\{q|\sum\nolimits_{\mathbf{y}:\ C(\bxy)=1} q(\byx)=r\},
        \end{equation*}
        where Eq. \eqref{eq:Q} is the special case when $r=1.$ The general token-level closed-form solution is
        $$q^*(y_i|\mathbf{x},\prefix{i})=\frac{\alpha R^C_p(\mathbf{x},\prefixeq{i})+\beta (1-R^C_p(\mathbf{x},\prefixeq{i}))}{\alpha R^C_p(\mathbf{x},\prefixeq{i-1}) +\beta(1-R^C_p(\mathbf{x},\prefixeq{i-1}))}p(y_i|\mathbf{x},\prefix{i}),$$
        where $\alpha=\frac{r}{R^C_p(\mathbf{x})}, \beta=\frac{1-r}{1-R^C_p(\mathbf{x})}.$
        
        Similar to Eq. \eqref{eq:toklvl}, once we have access to $R^C_p$, we can directly compute the closed-form solution even though the form is much more complicated. In this paper we only focus on hard constraints ($r=1$), however, here we demonstrate that our framework is capable of handling soft constraints as well. %for the relaxed controlling.
        
    \subsection{Approximating $R^C_p$ by NADO and Theoretical Analysis} \label{sec:bounds}
        Unfortunately, function $R^C_p$ defined in Eq. \eqref{eq:rc} is intractable. We cannot enumerate all possible sequences $\mathbf{y}$ since the space is exponentially large and essentially infinite. Hence, we train a neural model NADO to approximate this well-defined function. We use $R^C_\theta$ to denote NADO parameterized by $\theta$. 
        In this section, we derive bounds to provide a theoretical analysis about the correlation between errors in approximation and errors in corresponding sequence-level distribution. Generally, when $R^C_\theta$ approximates $R^C_p$ precisely enough, we have an upper bound for the sequence-level distribution discrepancy. The following lemma provides the formal definition.
        
        \noindent \textbf{Lemma 1} We define distribution
        \begin{equation}
        %\small
        \label{eq:q}
            q(y_i|\mathbf{x},\prefix{i})\propto\frac{R^C_\theta(\mathbf{x},\prefixeq{i})}{R^C_\theta(\mathbf{x},\prefixeq{i-1})}p(y_i|\mathbf{x},\prefix{i}).
        \end{equation}
        If there exists $\delta>1$ such that given input $\mathbf{x},$ $\forall \prefix i, $
        $\frac 1\delta < \frac{R^C_\theta(\mathbf{x},\prefixeq i)}{R^C_p(\mathbf{x}, \prefixeq i)} < \delta,$
        we have
        $$ KL(q^*(\byx)\|q(\byx))<(2L+2) \ln \delta,$$
        where $L$ is the length of the sequence $\mathbf{y}.$
        
        We also notice that by definition, $R^C_p$ satisfies the following equation:
        \begin{equation}
        %\small
        \label{eq:reg}
            \sum\nolimits_{y_i}R^C_p(\mathbf{x},\prefixeq{i})p(y_i|\mathbf{x},\prefix{i})=R^C_p(\mathbf{x},\prefixeq{i-1}).
        \end{equation}
        
        If $R$ also satisfies Eq. \eqref{eq:reg}, we can tighten this bound. Formally,
        
        \noindent \textbf{Lemma 2} Given the condition in Lemma 1, if $q$ is naturally a valid distribution without normalization (\emph{i.e.}, $\sum_{y_i} \frac{R^C_\theta(\mathbf{x},\prefixeq{i})}{R^C_\theta(\mathbf{x},\prefixeq{i-1})}p(y_i|\mathbf{x},\prefix{i})=1$), we have
         $$\forall x, KL(q^*(\byx)\|q(\byx))<2 \ln \delta.$$
        
        This lemma shows that with the auto-regressive property, the error does not accumulate along with the sequence. The proof is in the appendix. These two bounds indicate that when training the model $R^C_\theta$, we should push it to satisfy Eq. \eqref{eq:reg} while approximating $R^C_p$.
        
    \subsection{Training NADO}
    \label{sec:train}
        In Fig. \ref{fig:rc} we show the architecture of NADO. In general, NADO can be any seq2seq model. During training, it takes $\bxy$ as input and predicts from $R^C_\theta(\mathbf{x},\prefixeq 0)$ to $R^C_\theta(\mathbf{x},\prefixeq T).$ During the inference time, there are two parallel forward pass\footnote{In practice, to avoid enumerating the vocabulary, $R^C_\theta$ outputs a vector over vocabulary (\emph{i.e.}, $R^C_\theta(\mathbf{x}, \prefixeq {i-1} \oplus y)$ for all possible $y$, $\oplus$ is the concatenation operation), then we can directly do element-wise multiplication between $R^C_\theta$ and $p$.} to compute the token distribution $q$.  Considering the size of the NADO is usually much smaller than the base model, the whole forward pass takes no more than 2x base model forward pass time.
        % \begin{figure}[t]
        %     \centering
        %     \includegraphics[width=0.7\linewidth]{ConsDecIllustration.pdf}
        %     \caption{The architecture of model $R^C_\theta$. The base model, in each step, output a token distribution. Guided by $R^C_\theta$, we get distribution $q$, based on which we can generate the current token. $R^C_\theta$ is independently trained, and do jointly inference with the base model. Although the auxiliary model architecture is similar as base model, in practice, the auxiliary model can share the embedding part with base model, and use smaller decoder to reduce the number of parameters.}
        %     \label{fig:rc}
        % \end{figure}
        
        Now we discuss the training objective. In training, with some predefined input distribution $\mathcal{X},$ we sample $\mathbf{x}\sim \mathcal{X},$ $\mathbf{y}\sim p(\byx).$ We take these sampled $(\bxy)$ pairs as training examples, and use the boolean value $C(\bxy)$ as their labels for all steps. We use cross entropy (denoted as $CE(\cdot,\cdot)$) as the loss function, formally,
        $L_{CE}(\bxy,R^C_\theta)=\sum\nolimits_{i=0}^T CE(R^C_\theta(\mathbf{x},\prefixeq i), C(\bxy)).$
        Given a particular input $\mathbf{x}$, in expectation, we have
        \begin{equation}
        \label{eq:ExpCE}
        %\small
        \begin{aligned}
            \mathbb{E}_{ \mathbf{y}\sim p(\byx)}L_{CE}(\bxy,R^C_\theta) 
            &=\sum\nolimits_{\mathbf{y}\in\mathcal{Y}} p(\byx)L_{CE}(\bxy,R^C_\theta) \\
            &=\sum_{i=0}^T R^C_p(\mathbf{x},\prefixeq i) \!\log R^C_\theta(\mathbf{x},\prefixeq i) \!\!+\!\! (1\!\!-\!\!R^C_\theta(\mathbf{x},\prefixeq i))\log(1\!\!-\!\!R^C_\theta(\mathbf{x},\prefixeq i))\\
            &=\sum_{i=0}^T CE(R^C_p(\mathbf{x},\prefixeq i), R^C_\theta(\mathbf{x},\prefixeq i))\\
        \end{aligned}
        \end{equation}
        
        % \begin{equation}
        % \label{eq:ExpCE}
        %     \mathbb{E}_{ \mathbf{y}\sim p(\byx)}L_{CE}(\bxy,R^C_\theta) 
        %     =\sum_{\mathbf{y}\in\mathcal{Y}} p(\byx)L_{CE}(\bxy,R^C_\theta) 
        %     =\sum_{i=0}^T CE(R^C_p(\mathbf{x},\prefixeq i), R^C_\theta(\mathbf{x},\prefixeq i))\\
        % \end{equation}
        
        Therefore, $L_{CE}$ empirically estimates the cross entropy loss between $R^C_\theta$ and the ground truth $R^C_p$ which is intractable. 
        
        As we analyze above, we also regularize $R^C_\theta$ for satisfying Eq. \eqref{eq:reg} based on KL-divergence:
        \begin{equation*}
        %\small
            L_{reg}(\bxy,R^C_\theta)=f_{KL}\left(\sum_{y_i}R^C_\theta(\mathbf{x},\prefixeq{i})p(y_i|\mathbf{x},\prefix{i}),R^C_\theta(\mathbf{x},\prefixeq{i-1})\right).
        \end{equation*}
        
        $f_{KL}(p,q)=p\log \frac pq + (1-p)\log\frac{1-p}{1-q}$ is KL-divergence regarding $p$ and $q$ as two Bernoulli distributions. We use a hyper-parameter $\lambda>0$ to balance these losses. The final  training loss is
        \begin{equation}
        \label{eq:loss}
        %\small
            L(\bxy,R^C_\theta)=L_{CE}(\bxy,R^C_\theta)+\lambda L_{reg}(\bxy,R^C_\theta).
        \end{equation}
        
        %The whole algorithm can be describe as below: given base model $p$ and boolean oracle function $C$, we first sample $(\bxy)$ pairs from model $p$ and acquire the corresponding labels from $C$. With $((\bxy),C(\bxy))$ we train model $R^C_\theta$ with objective defined in Eq. \eqref{eq:loss}. This auxiliary model $R^C_\theta$ together with base model defines a regularized distribution $q$ defined in Eq. \eqref{eq:q}, based on which we are able to decode sequences that better satisfy boolean oracle function $C$.
        
    \subsection{Sampling}
    \label{sec:sampling}
        In Sec. \ref{sec:train} we describe that we train NADO by sampled data from base model $p$. One advantage is that we are able to leverage different sampling strategies to better adapt to different application scenarios. It is also possible to leverage reinforcement learning to train $R^C_\theta$, and we discuss our connection to reinforcement learning in the appendix. In this section, we introduce two sampling strategies and their corresponding properties.
        
        \noindent \textbf{Sampling with Temperature Control.}
        In some task, the output sequences are not diverse much, in other words, the token distribution in each step is very peaky. Since our NADO is trained on the sampled examples, we expect those examples to cover as much tokens combination as possible to avoid overfitting. Therefore, we add temperature factor $T$ to smooth the distribution \citep{ackley1985learning}. Specifically, we sample $\mathbf{y}$ from distribution $p(\byx)^{\frac{1}{T}}$, and add coefficient $p(\byx)^{1-\frac{1}{T}}$ when computing the cross-entropy loss. Formally, the expected loss is
        \begin{equation*}
        %\small
            \mathbb{E}_{\mathbf{y}\sim p(\byx)^\frac{1}{T}}\left[p(\byx)^{1-\frac{1}{T}}L_{CE}(\bxy,R^C_\theta)\right]
            =\sum\nolimits_{\mathbf{y}\in\mathcal{Y}} p(\byx)L_{CE}(\bxy,R^C_\theta),
        \end{equation*}
        which is same as the original expected loss in Eq. \eqref{eq:ExpCE}.
    
        \noindent \textbf{Importance Sampling.}
        In practice, the training process of NADO can be extraordinarily difficult when samples generated by the base model $p$ hardly satisfy $C$. \emph{i.e.} $\mathbb{E}_{\mathbf{y} \sim p(\mathbf{y}|\mathbf{x})} [p(C|\mathbf{x}, \mathbf{y})] \simeq 0$. %This significantly affects the sampling efficiency for training $\hat{R}_\theta$. 
        Hence, we introduce the importance sampling \citep{hammersley1954importance} to tackle this issue. Basically, we leverage existing partially trained $\hat{R}_\theta$ to form distribution $\hat{q}$. Although $\hat{R}_\theta$ is not well-trained, it is still able to provide positive guidance to produce samples satisfying $C$. Note that $\hat{q}$ does not have to be updated in each training epoch. With coefficient $\frac{p(\byx)}{\hat{q}(\byx)}$, the expected loss is same as the original expected loss:
        \begin{equation*} 
        %\small
            \mathbb{E}_{\mathbf{y}\sim \hat{q}(\byx)}\left[\frac{p(\byx)}{\hat{q}(\byx)}L_{CE}(\bxy,R^C_\theta)\right]
            =\sum\nolimits_{\mathbf{y}\in\mathcal{Y}} p(\byx)L_{CE}(\bxy,R^C_\theta).
        \end{equation*}
        % In our experiments, we generally exploit two distributions:
        % \begin{itemize}
        %     \item (Teacher Forcing Warm-up) \label{text:is-mle} An empirical distribution $p_{data}^C$: available if we have access to some golden references that already satisfy $C$. We maximize the likelihood of these samples on $q$, but only back-propagating the gradient to the parameters of $R_{\theta}$. %
        %     This is usually used for constraints with very low initial satisfaction rate \emph{i.e.} $\mathbb{E}_{\mathbf{y} \sim p(\mathbf{y}|\mathbf{x})} [p(C|\mathbf{x}, \mathbf{y})] \simeq 0$.
        %     %or we have only the unconditional distribution $p(\mathbf{y})$ and will have to fully rely on $R^C_\theta$ to impose the constraints.
        %     \item \label{text:is-q}($q$-Guided Importance Sampling) $\hat{q}(\mathbf{y}|\mathbf{x})$ using \emph{an existing approximation} $R_p^C$. Although $\hat{q}$ (as in Eq.~\ref{eq:q}) could still be sub-optimal, it is still a distribution more probably produces $C$-constrained samples compared to $p$. This is particularly useful after we have warmed-up $R_p^C$ with teacher forcing. Note that $\hat{q}$ does not have to be the up-to-date $q_\theta$. With coefficient $\frac{p(\byx)}{\hat{q}(\byx)}$, the expected loss is same as original expected loss:
        %     \begin{equation*} 
        %     \mathbb{E}_{\mathbf{y}\sim \hat{q}(\byx)}\left[\frac{p(\byx)}{\hat{q}(\byx)}L_{CE}(\bxy,R^C_\theta)\right]
        %     =\sum_{\mathbf{y}\in\mathcal{Y}} p(\byx)L_{CE}(\bxy,R^C_\theta),
        % \end{equation*}
        % \end{itemize}

%% file: experiment.tex
\section{Experiments}
    We conduct experiments on two tasks: lexically constrained generation (LCG) and machine translation (MT) with formality change. For the former, we use GPT-2 \citep{radford2019gpt2} as the base model and for the latter, we use a sequence-to-sequence model, MarianMT \citep{marcin2018marian}. We demonstrate our framework is generally effective in both scenarios. The boolean oracle is a rule-based function checking whether all lexical constraints are satisfied in LCG task, while in MT it is a classifier trained on an external dataset identifying the formality of the text. We put all details about hyper-parameter settings in the appendix.   
    
    \subsection{Text Generation with Lexical Constraints}
    We evaluate our model on two general classes of LCG problems: 
    \begin{itemize}
        \item Unsupervised LCG: annotation for lexical constraints are not available during training, but are expected to be in their exact order and lexical form during inference.
        \item Supervised LCG: annotation for lexical constraints are available, yet the words may appear in a different lexical form (e.g., ``look'' can appear in the past tense ``looked'') or a different order in the generated text.
    \end{itemize}
    %they could be re-organized in the generated contents in a more flexible way, like in arbitrary orders and tenses (e.g., ``look'').
    In both cases, we define oracle $C$ as a boolean function indicating whether the generated sequence \emph{satisfies all of the lexical constraints}. We do not naturally have negative samples (\emph{i.e.} the sequences that do not satisfy all constraints) to train the auxiliary model in both settings, thus, it is non-trivial to compare against methods requiring both positive and negative labeled data for training the auxiliary model like FUDGE and GeDi.
    
    \noindent \textbf{Data Setup} For unsupervised LCG, we follow the settings in POINTER \citep{zhang2020pointer} and conduct our experiments on Yelp! Review and News dataset. Each of the unsupervised LCG dataset contains a great number of un-annotated, raw sequences for training (160K for Yelp! Review and 268,586 for News). During inference, the model is expected to generate text lexically constrained in the exact order and form by a specific number of keywords (7 for Yelp! Review and 4 for News). For supervised LCG, we evaluate the proposed method on CommonGen \citep{lin2019comgen}. CommonGen is a supervised LCG task that aims to examine the commonsense of neural text generation models. For training, it contains 32,651 unique key \emph{concepts} (\emph{i.e.} the constraints) with 67,389 completed sequences in total. It also contains a validation set with 993 \emph{concepts} and 4018 reference sequences. For a more robust evaluation, the dataset maintains an open leaderboard that benchmarks different approaches on a withheld test set. We follow most of the data configurations specified in the original paper that first introduced the datasets.
    
    \noindent \textbf{General Model Setup} We investigate the effectiveness of different factors in our framework by enumerating different combinations of them. We implement two types of base model: 
    \begin{compactitem}
        \item (Seq2seq base model) A sequence-to-sequence model $p(\mathbf{y}|\mathbf{x})$ that takes into account the lexical constraints as condition sequence input;
        \item (DA base model) A language model that is only domain-adapted to $p(\mathbf{y})$ but unconditioned on anything. This is a challenging setting, since we impose the lexical constraints only with NADO. This setting is to better verify the effectiveness and efficiency of the proposed method and control irrelevant factors.
    \end{compactitem}
    Under both $p(\mathbf{y}|\mathbf{x})$ and $p(\mathbf{y})$ settings, we fine-tune the base model from the pre-trained GPT2-Large.
    
    During training, NADO is trained as a Seq2seq-like model\footnote{In this experiment, the input $\mathbf{x}$ is only describing the lexical constraint $C$. However, our framework also supports general inputs in other Seq2seq tasks with constraints. For example, machine translation with lexical constraints where the constraint $C$ is different from the input $\mathbf{x}$.}, which takes in the keys (for unsupervised LCGs, they are generated by randomly sampling a specific number of natural words in the original sentence) and generates the token-level guidance $R^C_\theta(\mathbf{x}, \mathbf{y}_{\leq i})$. For each pseudo key, we sample 32 target text with top-p ($p=0.8$) random sampling from base model $p$. 
    %For Seq2seq base models, we test setups with/without teacher-forcing based warmup (described in \ref{text:is-mle}). With DA base models, however, the warmup process is always incorporated for practical success of training. To verify that the major improvement of our framework is not coming from the extended capacity caused by the extra parameters $\theta$, we test setups with/without the proposed sample-oriented training (\emph{i.e.} the NADO objective as in Eq.~\ref{eq:loss}).
    We conduct experiments to test different training setups for NADO:
    \begin{compactitem}
        \item (NADO training)  The proposed training process described in Sec. \ref{sec:train}.
        \item (Warmup) We warm up NADO by maximizing the likelihood of positive samples, but only backpropagating the gradient to the parameters of $R_{\theta}$. The warm-up $R^C_\theta$ is used for importance sampling described in Sec. \ref{sec:sampling}. With DA base models, however, the warmup process is always incorporated for practical success of training (see the results for DA pretrained w/o warmup). 
    \end{compactitem}
    We also consider the setting with warmup only, which can be treated as a stronger baseline to verify that the major improvement of our framework is not coming from the extended capacity in NADO.
    \begin{figure}[t]
        \centering  
        \begin{subfigure}{0.4\linewidth}
            \includegraphics[height=0.7\linewidth]{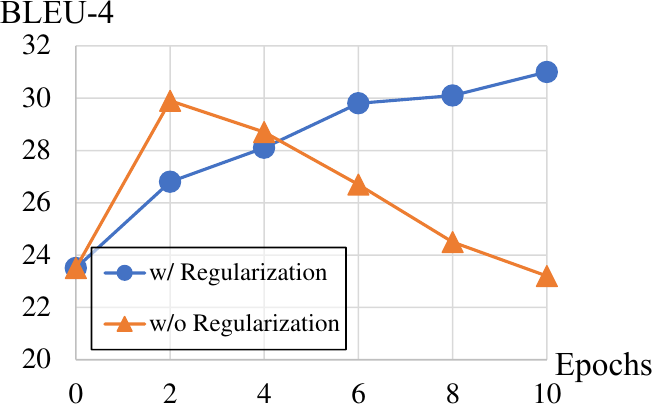}
            \caption{BLEU-4 comparison of NADO training with/without using Eq. \ref{eq:reg}}
            \label{fig:bleu4regimpact}
        \end{subfigure}
        \hspace{10pt}
        \begin{subfigure}{0.4\linewidth}
            \includegraphics[height=0.7\linewidth]{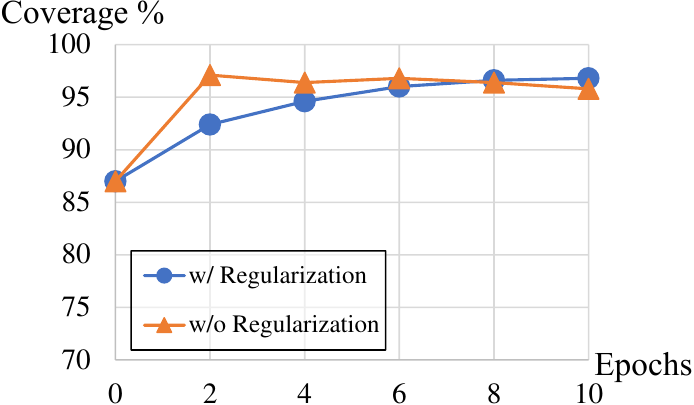}
            \caption{Coverage comparison of NADO training with/without using Eq. \eqref{eq:reg}.}
            \label{fig:coverageregimpact}
        \end{subfigure}
        \caption{Comparative study of the effectiveness of regularization in NADO training.}
        \label{fig:regimpact}
    \end{figure}

    \noindent \textbf{Results and Analysis} We compare the performance under different setups of our model to previous state-of-the-art methods on LCG tasks, including insertion-based models (Levenshtein Transformer \citep{gu2019levenshtein} with Lexical Constraints \citep{susanto2020lexicallevenshtein}, InsNet \citep{lu2021insnet}, etc.) and decoding-based algorithms. We also compare the results with a simple baseline which address the problems with a standard Seq2seq pipeline. The results are as shown in Table~\ref{tab:lcg}. 
    % NADO achieves significantly better generation quality compared to previous GPT-2-Large-based methods and even most insertion-based models of the same size. Compared to the fine-tuning baseline, the lexical coverage of NADO is significantly improved. 
    
    NADO consistently improves the BLEU score and coverage in different setups. Furthermore, under the best setting of each task (see bolded items in the table), NADO performs significantly better than most baselines in generation quality and can achieve very good lexical constraints coverage rate. Compared to InsNet, it is much easier for an autoregressive model with NADO to handle flexible reordering/transformation of lexical constraints. This is reflected in the performance comparison of InsNet and NADO on CommonGen dataset. Under most settings, a Seq2seq base model makes it easier for the framework to perform well, as it guarantees a reasonable level of lexical constraint coverage in even the initial state of the model. 
    
    \cready{Using a DA pretrained base model is a even challenging setup since the lexical constraints are only imposed with NADO. Therefore, the base model distribution is much distinct from the one filtered by the oracle, which is shown by poor performances on both metrics. However, with warmup and NADO under importance sampling, we show that it is still possible to obtain a powerful model with the proposed method. }
    
    \cready{To further study the correlation between the base model quality and the improvement of NADO, we conduct experiments on GPT-2 base model. The GPT-2 base model has lower scores with and without NADO compared with GPT-2 large, while the coverage improvements are similar. It shows NADO is capable to push the base model distribution towards the oracle if the base model has decent quality.}
    
    \cready{We also do human evaluation on base model (GPT-2 Large fine-tune) and the best NADO system, together with the gold reference for comparison. The results are shown in Tab. \ref{tab:human_lcg}. The evaluation metrics are detailed described in the Appendix. Some qualitative are shown in Tab. \ref{tab:twitter}.}
     
    To study the importance of the regularization term, we conduct an ablative study under the optimal setting on the CommonGen dataset (Seq2seq base model with NADO only). The results are shown in Figure~\ref{fig:regimpact}. While the success of achieving lexical control does not degenerate when NADO w/o regularization overfits, adding regularization can significantly improve the robustness of NADO generation quality when training NADO for more epochs.
    
    %\tao{1. NADO consistently improve the BLEU score and coverage in differnet setups. 2. Compared to previous work, We are significantly better in generation quality. Insertion based cannot handle word order. 3. Seq2Seq is better than DA, NADO is better than warmup + NADO, hypothesis: disentanglement benefit. 4. Regularization helps avoid overfitting. }
    
        \begin{table}[t]
            \renewcommand{\tabcolsep}{0.3mm}
            \centering
            \setlength{\abovecaptionskip}{10pt}
            \setlength{\belowcaptionskip}{10pt}
            \caption{Unsupervised/Supervised Lexically Constrained Generation results on Yelp Review (unsupervised), News (unsupervised) and CommonGen (supervised) dataset. CVRG stands for constraints coverage. For insertion-based models, on CommonGen dataset we directly use the keyword as initial context with no further permutation. $p, q$ denote the base model and the combined model in our framework, respectively. The domain adaptation pretrained model produces samples unconditioned on the constraints, and thus results in worse results than other setups. Results with * mark are from the open leader board on the test set instead of development set. }
            \resizebox{\textwidth}{!}{
            \begin{tabular}{l|ll|ll|ll}
                \toprule
                Dataset                     & \multicolumn{2}{l|}{Yelp Review (test)}                & \multicolumn{2}{l|}{News (test)}                       & \multicolumn{2}{l}{CommonGen (dev)}                   \\ \midrule
                Metrics                     & \multicolumn{1}{l|}{BLEU-2/4}   & CVRG & \multicolumn{1}{l|}{BLEU-2/4}   & CVRG & \multicolumn{1}{l|}{BLEU-3/4}    & CVRG \\\midrule
                \textbf{Insertion-based Baselines} & \multicolumn{1}{l|}{} &       & \multicolumn{1}{l|}{} &       & \multicolumn{1}{l|}{} & \\
                InsNet-Sequential \citep{lu2021insnet}        & \multicolumn{1}{l|}{19.4/5.8} & 100\%       & \multicolumn{1}{l|}{16.3/5.0} & 100\%       & \multicolumn{1}{l|}{26.2/18.7} & 100\%         \\ 
                
                ConstLevT \citep{susanto2020lexicallevenshtein}       & \multicolumn{1}{l|}{14.8/4.0} & 100\%       & \multicolumn{1}{l|}{11.8/1.9} & 100\%       & \multicolumn{1}{l|}{21.3/12.3*} & 96.9\%*         \\ 
                % InsT-POINTER-Large (w/ BERT init+Wiki) & \multicolumn{1}{l|}{16.8/3.8} & 100\%       & \multicolumn{1}{l|}{14.0/3.0} & 100\%       & \multicolumn{1}{l|}{-} & -         \\  
                \midrule
                \textbf{Algorithmic Baselines} & \multicolumn{1}{l|}{} &       & \multicolumn{1}{l|}{} &       & \multicolumn{1}{l|}{} & \\
                GPT-2-Large Finetune + Sampling       & \multicolumn{1}{l|}{16.4/5.3} & 94.5\%       & \multicolumn{1}{l|}{13.2/4.2} & 81.8\%       & \multicolumn{1}{l|}{34.2/24.7*} & 82.2\%*          \\ 
                Neural Logic \citep{lu2021insnet}               & \multicolumn{1}{l|}{-}          & -             & \multicolumn{1}{l|}{-}          & -             & \multicolumn{1}{l|}{36.7/26.7*}   & 97.7\%*         \\ 
                A*esque Decoding \citep{lu2022neurologic}               & \multicolumn{1}{l|}{-}          & -             & \multicolumn{1}{l|}{-}          & -             & \multicolumn{1}{l|}{-/28.2*} &   97.6\%*    \\ 
                \midrule
                \textbf{Model Setups (Ours)} & \multicolumn{1}{l|}{} &       & \multicolumn{1}{l|}{} &      & \multicolumn{1}{l|}{} &           \\ 
                $p$ (Domain Adaptation pretrain) & \multicolumn{1}{l|}{5.3/0.4} & 5.4\%       & \multicolumn{1}{l|}{4.0/0.8} & 0.9\%       & \multicolumn{1}{l|}{9.3/3.9} & 8.5\%           \\ 
                $p$ (Seq2seq pretrain) & \multicolumn{1}{l|}{16.6/4.8} & 91.2\%       & \multicolumn{1}{l|}{13.0/3.4} & 74.0\%       & \multicolumn{1}{l|}{34.2/23.5} & 87.0\%           \\ 
                $q$ (DA pretrained $p$ + warmup) & \multicolumn{1}{l|}{16.2/4.3} & 75.4\%       & \multicolumn{1}{l|}{12.6/2.8} & 66.7\%       & \multicolumn{1}{l|}{32.7/20.9} & 79.7\%           \\ 
                $q$ (DA pretrained $p$ + warmup + NADO) & \multicolumn{1}{l|}{16.9/5.4} & 95.6\%       & \multicolumn{1}{l|}{\textbf{15.4/4.7}} & \textbf{92.3\%}       & \multicolumn{1}{l|}{37.8/26.2} & 96.1\%           \\  
                $q$ (Seq2seq pretrained $p$ + warmup) & \multicolumn{1}{l|}{16.8/5.7} & 94.2\%       & \multicolumn{1}{l|}{13.6/4.2} & 85.0\%       & \multicolumn{1}{l|}{35.2/24.8} & 90.2\%           \\ 
                $q$ (Seq2seq pretrained $p$ + NADO) &  \multicolumn{1}{l|}{\textbf{17.4/6.0}} & \textbf{96.7\%}       & \multicolumn{1}{l|}{15.0/4.5} & 91.9\%       & \multicolumn{1}{l|}{\textbf{40.9/30.8}} & \textbf{97.1\%}           \\ 
                $q$ (Seq2seq pretrained $p$ + warmup + NADO) & \multicolumn{1}{l|}{16.7/4.7} & 92.8\%       & \multicolumn{1}{l|}{14.4/4.4} & 86.1\%       & \multicolumn{1}{l|}{40.2/30.3} & 95.9\%           \\ \midrule
                \textbf{GPT-2 Base Reference} & \multicolumn{1}{l|}{} &       & \multicolumn{1}{l|}{} &      & \multicolumn{1}{l|}{} &           \\ 
                $q$ (Seq2seq pretrained $p$) & \multicolumn{1}{l|}{-} & -       & \multicolumn{1}{l|}{-} & -      & \multicolumn{1}{l|}{32.17/22.98} & 76.8\%           \\ 
                $q$ (Seq2seq pretrained $p$ + NADO) & \multicolumn{1}{l|}{-} & -       & \multicolumn{1}{l|}{-} & -       & \multicolumn{1}{l|}{33.61/24.01} & 85.5\%           \\
                \bottomrule
            
            \end{tabular}}
            
            \label{tab:lcg}
        \end{table}
        
        % \begin{table}[t]
        %     \centering
        %     \begin{tabular}{c|c|c|c|c}
        %          Model & Quality & Plausibility & Concepts & Overall \\
        %          \midrule
        %          Baseline & 2.50 & 2.55 & 2.59 & 2.47 \\
        %          NADO (Ours) & \textbf{2.65} & 2.61 & \textbf{2.75} & 2.54 \\
        %          Gold Ref. & 2.62 & \textbf{2.65} & 2.66 & \textbf{2.64} \\
        %     \end{tabular}
        %     \caption{Human evaluation of generated texts in CommonGen test set. The detailed description for the four metrics (scale: from 1 to 3) can be found in the Appendix. Baseline stnads for GPT-2 Large fine-tune setting, and NADO stands for the best system, Seq2seq pretrained + NADO. We also evaluate the first gold reference provided in the dataset for comparison.}
        %     \label{tab:human_lcg}
        % \end{table}
        
        \begin{table}[t]
            \centering
            \caption{Human evaluation of generated texts in CommonGen test set. The detailed description for the four metrics (scale: from 1 to 3) and the evaluation setups can be found in the Appendix. Baseline stnads for GPT-2 Large fine-tune setting, and NADO stands for the best system, Seq2seq pretrained + NADO. We also evaluate the first gold reference provided in the dataset for comparison. NADO outperforms base model in all four metrics. (The difference is statistical significant tested by Wilcoxon signed ranks one-sided test, $p$-value < 0.02) }
            \vspace{0.1in}
            \begin{tabular}{c|c|c|c|c}
                 \toprule
                 Model & Quality & Plausibility & Concepts & Overall \\
                 \midrule
                 Baseline & 2.39 & 2.46 & 2.40 & 2.37 \\
                 NADO (Ours) & 2.51 & 2.52 & 2.52 & 2.47 \\
                 Gold Ref. & 2.53 & 2.58 & 2.59 & 2.56 \\
                 \bottomrule
            \end{tabular}
            
            \label{tab:human_lcg}
        \end{table}
        
        \begin{table}[t]
            \small
            \centering
            \setlength{\abovecaptionskip}{10pt}
            \setlength{\belowcaptionskip}{0pt}
            \caption{Some more qualitative generation results with randomly selected concepts about NeurIPS.}
            \begin{tabularx}{\textwidth}{lXX}
            \toprule
                Constraint: & The generated texts should contain all the given concepts in arbitrary order \\
            \midrule
            \midrule
                Concepts & \rb{look} \bb{forward} \gb{discuss} \pb{NeurIPS}\\
                Base Model Sample \#1 & Players \gb{discuss} the \rb{look} of \bb{forward} NeurrIPS. (\pb{\sout{NeurIPS}}) \\
                Base Model Sample \#2 & Football player and \bb{forward} \gb{discuss} a \rb{look} at the move. (\pb{\sout{NeurIPS}})\\
                NADO Sample \#1 & People \rb{look} \bb{forward} to \gb{discussing} the future of \pb{NeurIPS}. \\
                NADO Sample \#2 & We \rb{look} \bb{forward} to meeting and \gb{discussing} the future of \pb{NeurIPS}. \\
                
            \end{tabularx}

            \begin{tabularx}{\textwidth}{lXX}
            \midrule\midrule
                Concepts & \rb{excite} \bb{paper} \gb{accept}  \pb{NeurIPS} \\
                Base Model Sample \#1 & Researchers are \rb{excited} after \gb{acceptance} of their \bb{paper} at IPS. (\pb{\sout{NeurIPS}}) \\
                Base Model Sample \#2 & Scientists \rb{excited} to \gb{accept} \bb{paper} \gb{accepted} at \pb{NeurIPS}. \\
                NADO Sample \#1 & \pb{NeurIPS} is \rb{excited} to \gb{accept} the \bb{paper} of researcher. \\
                NADO Sample \#2 & \pb{NeurIPS} is \rb{excited} to announce that it has \gb{accepted} \bb{papers}. \\
            \bottomrule
            \end{tabularx}

        \label{tab:twitter}
    \end{table}

    \subsection{Machine Translation with  Formality Change}
        
        \paragraph{Datasets and  Setup}
        We follow the experimental setting in FUDGE \citep{yang2021fudge} to formalize the results of machine translation. Given an informal source sentence, our goal is to translate it into formal sentence written in the target language. We conduct our experiments on Fisher and CALLHOME Spanish-English Speech Translation Corpus \citep{post2013fisher}, where both of the Spanish source and English reference are informal and casual. Instead of evaluating the translation on original references, we use the formal and fluent rewritten version of references \citep{salesky2019fluent} to evaluate the translation quality by BLEU scores. In the training process, the formal version reference is unseen to the models. We also evaluate the formality scores by a discriminator trained on GYAFC formality dataset
        \citep{rao2018dear} as what FUDGE paper does. In this experiment, pre-trained Marian MT model \citep{marcin2018marian} is used as the base model. 
        
        In FUDGE, the authors train an auxiliary model also on GYAFC modeling token-level guidance $P(\mbox{formal}|\prefix i)$, and leverage it to guide the base model by Bayesian rule 
        \begin{equation}
        \label{eq:bayesian}
            P(y_i|\prefix i, \mbox{formal})\propto P(y_i|\prefix i) P(\mbox{formal}|\mathbf{y}_{\leq i}).
        \end{equation}

        For the formality supervision, FUDGE leverages an external token-level oracle. In NADO, we load the same oracle but exclusively leverage sequence-level binary supervision as oracle $C$. We randomly choose 10,000 (7.2\%) source texts from the training set as input examples, and sample $8$ target texts by sampling with temperature $T$ from base model $p$ for each source text.  We use those sampled examples to train NADO. In total, we have $80,000$ training samples, which is similar to the number of training data (105k) for the token-level oracle in FUDGE. All the methods are using greedy decoding. 
        
        \noindent \textbf{Results and Discussion}
        The experimental results are shown in Table \ref{tab:mt}. Compared to FUDGE, although only the sequence-level supervision is leveraged, we are consistently better in both metrics, especially in BLEU score we boost about 3 points. We conjecture that the improvement is because our formulation is more principle and correct. In methods using auxiliary model to guide the base model, including FUDGE, their formulation is based on Eq. \ref{eq:bayesian}. However, the auxiliary model is trained on a distribution different from where the base model is pretrained on, which leads to a distributional discrepancy issue. In other words, directly multiplying these two terms is not rigorous, since they are estimated on two different distributions. On the contrary, NADO is trained specifically to the base model. This avoids the discrepancy issue and provides an accurate guidance. Considering we are using the same oracle function and similar number of training samples, the higher generation quality reflected by BLEU scores supports our conjecture.
        %which may have a distributional gap from where base model is pre-trained on. However, FUDGE directly combine them by incorrectly applying Bayesian rule, and this distribution discrepancy affects the performance. In NADO, the token-level guidance depends on the base model since it is trained by examples sampled from base model. Hence, our token-level guidance is aligned with the base model, resulting in higher translation quality.
        
        %We also conduct experiments on the effect of sampling with temperature. As we describe in Sec. \ref{sec:sampling}, sampling with temperature makes the generated sequences more diverse. In this MT task we sample 8 times from the each input, and we usually get 8 same translated texts by random sampling. Applying sampling by temperature is helpful for improving the coverage of different tokens in the sampled data, which is used to train model $R^C_\theta$, and further improve the quality of $R^C_\theta$ with proper temperature. 
        
        In sampling, for each input we sample 8 examples to train $R^C_\theta$, which are usually identical in this task. Applying temperature in sampling allows NADO to be trained with more diverse data. Results show that with a properly set temperature, we can further improve the generation quality.
        
        \cready{It is still possible that the neural oracle leverages some superficial or even spurious features and NADO is catering those features in order to improve the formality scores. For example, some informal little words like “hmm” “uh”, and some abbreviations like “ ‘cause ” “gonna” could make the formality score lower. We find that NADO tends to fix them (see Appendix D). However, how to get an good oracle is orthogonal to our contributions.}
         
        \begin{table}[t]
            \centering
            \setlength{\abovecaptionskip}{10pt}
            \setlength{\belowcaptionskip}{0pt}
            \caption{Formal Machine Translation results. We follow \citep{yang2021fudge} setting to choose BLEU score and average formality scores as the metric. We slightly improve the formality score compared to FUDGE, while significantly boost the BLEU score.}
            
            \begin{tabular}{c|c|c}
            \toprule
                Method & BLEU & Avg. Formality \\
                \midrule
                MarianMT \citep{marcin2018marian} & 16.98 & 0.45 \\
                FUDGE \citep{yang2021fudge} & 17.96 & 0.51 \\
                NADO + Random Sampling & 20.84 & 0.54 \\
                NADO + Sampling with $T=5/4$ & 21.04 & 0.53 \\
                NADO + Sampling with $T=5/3$ & 20.77 & 0.52 \\ \bottomrule
            \end{tabular}
            \label{tab:mt}
        \end{table}

%% file: checklist.tex
\section*{Checklist}
\begin{enumerate}

\item For all authors...
\begin{enumerate}
  \item Do the main claims made in the abstract and introduction accurately reflect the paper's contributions and scope?
    \answerYes{}
  \item Did you describe the limitations of your work?
    \answerYes{We emphasize the oracle should be boolean.}
  \item Did you discuss any potential negative societal impacts of your work?
    \answerYes{}
  \item Have you read the ethics review guidelines and ensured that your paper conforms to them?
    \answerYes{}
\end{enumerate}

\item If you are including theoretical results...
\begin{enumerate}
  \item Did you state the full set of assumptions of all theoretical results?
    \answerYes{See Sec. \ref{sec:bounds}}
        \item Did you include complete proofs of all theoretical results?
    \answerYes{In the appendix}
\end{enumerate}

\item If you ran experiments...
\begin{enumerate}
  \item Did you include the code, data, and instructions needed to reproduce the main experimental results (either in the supplemental material or as a URL)?
    \answerYes{Data is described the appendix and code will be uploaded in supplementary materials.}
  \item Did you specify all the training details (e.g., data splits, hyperparameters, how they were chosen)?
    \answerYes{In the appendix}
        \item Did you report error bars (e.g., with respect to the random seed after running experiments multiple times)?
    \answerNo{Generation is too computational expensive}
        \item Did you include the total amount of compute and the type of resources used (e.g., type of GPUs, internal cluster, or cloud provider)?
    \answerYes{In the appendix}
\end{enumerate}

\item If you are using existing assets (e.g., code, data, models) or curating/releasing new assets...
\begin{enumerate}
  \item If your work uses existing assets, did you cite the creators?
    \answerYes{}
  \item Did you mention the license of the assets?
    \answerYes{In the appendix}
  \item Did you include any new assets either in the supplemental material or as a URL?
    \answerYes{}
  \item Did you discuss whether and how consent was obtained from people whose data you're using/curating?
    \answerYes{In the appendix}
  \item Did you discuss whether the data you are using/curating contains personally identifiable information or offensive content?
    \answerNA{}
\end{enumerate}

\item If you used crowdsourcing or conducted research with human subjects...
\begin{enumerate}
  \item Did you include the full text of instructions given to participants and screenshots, if applicable?
    \answerNA{}
  \item Did you describe any potential participant risks, with links to Institutional Review Board (IRB) approvals, if applicable?
    \answerNA{}
  \item Did you include the estimated hourly wage paid to participants and the total amount spent on participant compensation?
    \answerNA{}
\end{enumerate}

\end{enumerate}

%% file: appendix.tex
\newpage
\section{Closed-form Token-level Decomposition}
    \subsection{Decomposition with Hard Constraints}
    The sequence-level solution $q^*$ is given by
    $$q^*(\byx)=\frac{p(\byx)C(\bxy)}{R^C_p(\mathbf{x})}.$$
    Now we prove that 
    $$q^*(y_i|\mathbf{x},\prefix{i})=\frac{R^C_p(\mathbf{x}, \prefixeq{i})}{R^C_p(\mathbf{x}, \prefixeq{i-1})}p(y_i|\mathbf{x},\prefix{i}),$$
    is the unique token-level decomposition.
    On one hand, we verify $q^*$ is a valid decomposition, which can be demonstrated by
    \begin{equation}
        \begin{aligned}
            \prod_{i=0}^L q^*(y_i|\mathbf{x},\prefix{i}) 
            &=\prod_{i=1}^L \frac{R^C_p(\mathbf{x}, \prefixeq{i})}{R^C_p(\mathbf{x}, \prefixeq{i-1})}p(y_i|\mathbf{x},\prefix{i}) \\
            &=\frac{R^C_p(\mathbf{x}, \prefixeq L)}{R^C_p(\mathbf{x}, \prefixeq 0)}\prod_{i=0}^L p(y_i|\mathbf{x},\prefix{i}) \\
            &=\frac{C(\bxy)}{R^C_p(\mathbf{x})}p(\byx)\\
            &=q^*(\byx),
        \end{aligned}
    \end{equation}
    together with
    \begin{equation}
        \sum_{y_i} q^*(y_i|\mathbf{x},\prefix{i})=\frac{\sum_{y_i}R^C_p(\mathbf{x}, \prefixeq{i})p(y_i|\mathbf{x},\prefix{i})}{R^C_p(\mathbf{x}, \prefixeq{i-1})}=1
    \end{equation}
    
    On the other hand, we demonstrate that the decomposition is unique. We generally prove that 
    \paragraph{Lemma 3} For finite space $\mathcal{X}\times \mathcal{Y}$, if the sequence-level distribution $q(\byx)$ is determined, the token-level distribution is unique. 
    \paragraph{Proof} We assume the longest sequence in $\mathcal{Y}$ has length $L$, and pad all the sequence to length $L$ by adding special token at the end. We prove the following statement by induction:
    
    Given input $\mathbf{x}$ and prefix length $i$, If $q(y_i|\mathbf{x},\prefix i)$ is determined, there exists a unique distribution $q(y_{i-1}|\mathbf{x},\prefix {(i-1)}).$
    
    When $i=L$, it is true since $y_i\oplus \prefix i$ is the full sequence ($\oplus$ denotes the string concatenation), hence $q(y_i|\mathbf{x},\prefix i)\propto q(y_i\oplus \prefix t|\mathbf{x}).$ $q(y_i\oplus \prefix t|\mathbf{x})$ is the determined sequence-level distribution, so $q(y_i|\mathbf{x},\prefix i)$ is unique.
    
    Assume the statement holds when $i>=t$, we consider $i=t-1.$ The distribution of prefixes with length $t-1$ is given by
    $$q(\prefix {(t-1)}|\mathbf{x})\propto \frac{q(\byx)}{\prod_{j=t}^L q(y_j|\mathbf{x},\prefix j)}.$$
    Since the sequence distribution and token-level distribution after step $t$ are determined, the prefix distribution is also determined. Thus, the token-level distribution at step $(t-1)$ has unique solution
    $$q(y_{t-1}|\mathbf{x},\prefix {(t-1)})\propto q(\prefix {(t-1)}|\mathbf{x}).$$
    
    \subsection{Closed-form Solution and Decomposition with Soft Constraints}
    To deal with soft constraints, we define the feasible set $Q$ as
    $$Q:=\{q|\sum\nolimits_{\mathbf{y}:\ C(\bxy)=0} q(\byx)=r\}.$$
    The sequence-level solution is given b
    \begin{equation*}
        \begin{aligned}
            q^*(\byx) 
            &=\arg\min_{q\in Q} KL(p(\byx)\|q(\byx))\\
            &=\arg\min_{q\in Q} \left[\sum_{\mathbf{y}:C(\bxy)=0} p(\byx)\log \frac{q(\byx)}{p(\byx)}+\sum_{\mathbf{y}:C(\bxy)=1} p(\byx)\log \frac{q(\byx)}{p(\byx)}\right]\\
        \end{aligned}
    \end{equation*}
    The optimal distribution $q^*$ should be proportional to $p$ in both term, respectively. Since $\sum_{\mathbf{y}:C(\bxy)=1} p(\byx)=R^C_p(\mathbf{x}),$ we have
    \begin{equation*}
        q^*(\byx)=
        \begin{cases}
            \frac{1-r}{1-R^C_p(\mathbf{x})}p(\byx) & C(\bxy)=0 \\
            \frac{r}{R^C_p(\mathbf{x})}p(\byx) & C(\bxy)=1 \\
        \end{cases}
    \end{equation*}
    We denote $\alpha=\frac{r}{R^C_p(\mathbf{x})},\ \beta=\frac{1-r}{1-R^C_p(\mathbf{x})}p(\byx).$ Now the sequence-level optimal distribution is determined, by Lemma 3, we have a unique token-level decomposition. Now we verify
    $$q^*(y_i|\mathbf{x},\prefix{i})=\frac{\alpha R^C_p(\mathbf{x}, \prefixeq{i})+\beta (1-R^C_p(\mathbf{x}, \prefixeq{i}))}{\alpha R^C_p(\mathbf{x}, \prefixeq{i-1}) +\beta(1-R^C_p(\mathbf{x}, \prefixeq{i-1}))}p(y_i|\mathbf{x},\prefix{i})$$
    is exactly what we want. We have

    \begin{equation*}
        \begin{aligned}
            \prod_{i=1}^L q^*(y_i|\mathbf{x},\prefix i)
            &=\prod_{i=1}^L \frac{\alpha R^C_p(\mathbf{x}, \prefixeq{i})+\beta (1-R^C_p(\mathbf{x}, \prefixeq{i}))}{\alpha R^C_p(\mathbf{x}, \prefixeq{i-1}) +\beta(1-R^C_p(\mathbf{x}, \prefixeq{i-1}))}p(y_i|\mathbf{x},\prefix{i}) \\
            &=\frac{\alpha R^C_p(\mathbf{x}, \prefixeq L)+\beta(1-R^C_p(\mathbf{x}, \prefixeq L))}{\alpha R^C_p(\mathbf{x})+\beta(1-R^C_p(\mathbf{x}))}\prod_{i=1}^L p(y_i|\mathbf{x},\prefix i) \\
            &=\frac{\beta+(\alpha-\beta)C(\bxy)}{r+(1-r)}p(\byx) \\
            &=q^*(\byx),
        \end{aligned}
    \end{equation*}
    and
    $$\sum_{y_i} q^*(y_i|\mathbf{x},\prefix i)=1,$$
    since $$\sum_{y_i} R^C_p(\mathbf{x}, \prefixeq i))p(y_i|\mathbf{x},\prefix i)=R^C_p(\mathbf{x},  \prefixeq{i-1}).$$
    Therefore, $q^*$ is the unique solution.
    
\section{Proof of the Error Bounds}
    Here we prove the bounds in Lemma 1 and Lemma 2. \footnote{There are typos in the main text. Here we correct them. The typos do not affect related conclusions.}
    \paragraph{Proof of Lemma 1}
        We denote $Z_i(\bxy)$ as the normalization term in distribution $q(y_i|\mathbf{x},\prefix{i})$, \emph{i.e.},
        $$q(y_i|\mathbf{x},\prefix{i})=\frac{1}{Z_i(\bxy)}\frac{R^C_\theta(\mathbf{x}, \prefixeq{i})}{R^C_\theta(\mathbf{x}, \prefixeq{i-1})}p(y_i|\mathbf{x},\prefix{i}),$$
        $$Z_i(\bxy)=\sum_{y_i}\frac{R^C_\theta(\mathbf{x}, \prefixeq{i})}{R^C_\theta(\mathbf{x}, \prefixeq{i-1})}p(y_i|\mathbf{x},\prefix{i}).$$
        Since $\frac{1}{\delta}<\frac{R^C_\theta (\mathbf{x},  \prefixeq i)}{R^C_p (\mathbf{x},  \prefixeq i)}<\delta,$
        $$Z_i(\bxy)<\sum_{y_i}\delta^2\frac{R^C_p(\mathbf{x}, \prefixeq{i})}{R^C_p(\mathbf{x}, \prefixeq{i-1})}p(y_i|\mathbf{x},\prefix{i})=\delta^2.$$
        Similarly, $Z_i(\bxy)>\frac{1}{\delta^2}.$
        
        The sequence-level distribution is given by
        $$q(\byx)=\prod_{i=1}^L q(y_i|\mathbf{x},\prefix{i})=\frac{1}{\prod_{i=1}^L Z_i(\bxy)}\frac{R^C_\theta(\mathbf{x},  \prefixeq L)}{R^C_\theta(\mathbf{x})}p(\byx).$$
        Now we bound the KL-divergence as
        \begin{equation*}
            \begin{aligned}
                    KL(q^*(\byx)\|q(\byx))
                    &=\sum_{\mathbf{y}\in \mathcal{Y}} q^*(\byx)\log\left(\frac{q(\byx)}{q^*(\byx)}\right) \\
                    &=\sum_{\mathbf{y}:C(\bxy)=1} q^*(\byx) \log\left(\frac{1}{\prod_{i=1}^L Z_i(\bxy)}\frac{R^C_\theta(\mathbf{x})}{R^C_p(\mathbf{x} )}\frac{C(\bxy)}{R^C_\theta(\mathbf{x},  \prefixeq L)}\right)\\
                    &< \sum_{\mathbf{y}:C(\bxy)=1} q^*(\byx) \log (\delta^{(2L+2)})\\
                    &< (2L+2)\log \delta.
            \end{aligned}
        \end{equation*}
    \paragraph{Proof of Lemma 2}
        With additional condition, $Z_i(\bxy)=1.$. Hence the bound is $2\log\delta.$

\section{Experiment Details}
    \subsection{Datasets}
        In MT formality change experiment, we use Fisher and Callhome Spanish-English translation dataset \citep{post2013fisher} and GYAFC formality corpus \citep{rao2018dear}. Both of the datasets are not public. Fisher and Callhome dataset is owned by LDC. To acquire the GYAFC dataset, one need to first gain access to Yahoo Answers corpus.
        
        For unsupervised LCG experiments, we use Yelp Reviews \citep{cho2018towards} and WMT News section datasets \citep{bojar2017wmt,guo2018long}. Yelp Reviews is published under its own license \footnote{\hyperlink{https://s3-media0.fl.yelpcdn.com/assets/srv0/engineering_pages/dc1cabe7cb95/assets/vendor/Dataset_User_Agreement.pdf}{See full license file here.}}. Please refer to the official website of WMT dataset \citep{bojar2017wmt} for more information about licenses and legal concerns. We conduct our experiment of supervised LCG on the CommonGen dataset \citep{lin2019comgen}, whose official repository is published under MIT license.
    
    \subsection{Hyperparameters and Training Details}
        For MT experiments, we load the MarianMT from the es-en checkpoint provided by huggingface. We set a constant learning rate $1e-5$. The NADO shares the encoder with MarianMT, and use MarianDecoder as the decoding architecture. Compared to the 12-layer decoder in MarianMT, we a use 3-layer decoder, and the other configurations follow the Marian decoder. The fully connected output layer is zero initialized. We set the coefficient $\lambda=0.1$ for the regularization, which is selected from $\{0.01,0.03,0.1,0.3,1.0\}.$ All the hyperparameters are tuned on the development set. We select the best NADO model evaluated on development set in 10-epoch training.
    
        For LCG experiments, we finetune our base models from a GPT-2-Large checkpoint \citep{radford2019gpt2} for 3 epochs with learning rate $1e-5$. We use a warmup number of $400$ and the learning rate decays to 0 in $5000000$ steps (which are far more than the actually executed ones). We use NADO to train the auxiliary $R$ models (4-layer, 768-D, 12 headed transformer) with a learning rate of $2e-5$, which is selected from $\{1e-5, 2e-5, 5e-5\}$. We zero-initialize the output layer of auxiliary models. No learning warmup/decay is applied when training $R$. With the help of KL-regularization, we set $\lambda=1.0$ and we don't observe very severe overfitting in the second stage of NADO training. We simply report the results after the maximum number of training epochs (usually 20).
    
        For more implementation details and tricks, please refer to our code. \footnote{https://github.com/MtSomeThree/constrDecoding}
    \subsection{Computational Resources}
        We run the MT experiments on a NVIDIA GeForce RTX 2080 Ti GPU. We run LCG experiments on a NVIDIA A6000 GPU.
        
\section{Qualitative Generation Results}
    \subsection{Machine Translation Formality Change Experiments}
        In Tab.\ref{tab:MTcases} we show some translation text in the best system (NADO+Sampling with $T=5/4$) and baseline (MarianMT), together with the source and its formal reference.
        
        \begin{table}[t]
            \centering
            \setlength{\abovecaptionskip}{10pt}
            \setlength{\belowcaptionskip}{0pt}
            \caption{Some qualitative generation results in machine translation formality change experiments. NADO controls the model to be more formal by skipping informal words, fixing the grammar, capitalizing the first letters in base model generation results, etc. Noting that the zero-shot base model can not handle the repeated words in casual source language very well, we observe that the NADO-controlled model, which follows the base model distribution, cannot handle it, either. }
            \begin{tabularx}{\textwidth}{lXX}
            \toprule
                Constraint: & The generated (translated) texts should be fluent and formal.\\
            \midrule
            \midrule
                Source & Lo, y las eh usted dice que usted toca instrumentos, eh es- ¿usted su esposo tocan como en una banda, un grupo? \\
                MarianMT & \rb{What}, and \rb{the uh, you say} you play instruments, uh, you're-- you're your husband playing like a band, a band? \\
                NADO & \rb{So}, and \rb{you're saying that} you play instruments, uh, you're-- you're your husband playing like \rb{in} a band, a band? \\
                Formal Ref. & So, you say you play instruments, you and your husband play in a band or group? \\
            \end{tabularx}
        
            \begin{tabularx}{\textwidth}{lXX}
            \midrule\midrule
                Source & Entonces, ya los programas ya, ya están clasificados, ya ver una escena de de pornografía en la televisión a cualquier hora del día es normal. Y ya para muchos \\
                MarianMT & So, \rb{already the programs are already classified}, and seeing a porn scene on TV at any time of the day is normal. \rb{And already for many} \\
                NADO & So, \rb{the programs already, they're already classified}, and seeing a porn scene on TV at any time of the day is normal. \\
                Formal Ref. & The programs, are classified, to see a scene of pornography on the television at anytime of the day is normal. And for many \\
            \end{tabularx}
            
            \begin{tabularx}{\textwidth}{lXX}
            \midrule\midrule
                Source & donde hay problemas \\
                MarianMT & \rb{where} there's trouble. \\
                NADO & \rb{Where} there's trouble. \\
                Formal Ref. & That's when trouble arises. \\
            \bottomrule
            \end{tabularx}
        \label{tab:MTcases}
        \end{table}
    
    \subsection{Lexical Constrained Generation Experiments}
        In Tab.\ref{tab:MTcases} we show some generated text in best systems and baselines, together with the golden references.
         \begin{table}[t]
            \centering
            \setlength{\abovecaptionskip}{10pt}
            \setlength{\belowcaptionskip}{0pt}
            \caption{Some qualitative generation results in LCG benchmark experiments. NADO focuses on making sure all the constraints are imposed, or at least ensuring as many as possible of them. This is in particular more obvious when using weaker base models (e.g. GPT-2-base). }
            \begin{tabularx}{\textwidth}{lXX}
            \toprule
                Constraint: & The generated texts should contain all the given concepts in arbitrary order \\
            \midrule
            \midrule
                Concepts & \rb{kid} \rb{room} \rb{dance}  \\
                Base Model & A boy and girl \rb{dancing} in a \rb{room}. (\rb{\sout{kid}}) \\
                NADO & A \rb{kid} is \rb{dancing} in the \rb{room}. \\
                Golden Ref. \#1 & The silly \rb{kid} loves to \rb{dance} in her \rb{room}. \\
                Golden Ref. \#2 & the \rb{dance} \rb{kid}: \rb{room} is full of \rb{kids} \\
                Golden Ref. \#3 & A \rb{kid} is \rb{dancing} in the \rb{room}.\\
                Golden Ref. \#4 & A group of \rb{kids} are \rb{dancing} around a living \rb{room}. \\
            \end{tabularx}

            \begin{tabularx}{\textwidth}{lXX}
            \midrule\midrule
                Concepts & \rb{create} \rb{pottery} \rb{wheel}  \\
                Base Model & add a \rb{pottery} \rb{wheel} to your home. (\rb{\sout{create}}) \\
                NADO & \rb{create} a \rb{pottery} \rb{wheel} in the garden \\
                Golden Ref. \#1 & \rb{Create} \rb{pottery} with a \rb{wheel} and clay. \\
                Golden Ref. \#2 & I \rb{create} \rb{pottery} on a \rb{wheel}. \\
                Golden Ref. \#3 & A \rb{pottery} \rb{wheel} can be used to \rb{create} bowls.\\
                Golden Ref. \#4 & A man is using a \rb{wheel} to \rb{create} \rb{pottery}. \\
            \end{tabularx}

            \begin{tabularx}{\textwidth}{lXX}
                \midrule 
                \midrule
                Concepts & \rb{bed} \rb{look} \rb{sit}  \\
                Base Model & A man is \rb{sitting} on a \rb{bed}. (\rb{\sout{look}}) \\
                NADO & A man \rb{sits} on a \rb{bed} and \rb{looks} at his reflection in the mirror. \\
                Golden Ref. \#1 & The girl was \rb{sitting} in the couch and \rb{looking} at a bed in a catalog. \\
                Golden Ref. \#2 & the woman \rb{look} at her \rb{bed} as she \rb{sits}. \\
                Golden Ref. \#3 & The man \rb{sat} on the \rb{bed} and \rb{looked} out the window\\
                Golden Ref. \#4 & \rb{Looking} dejected, someone \rb{sits} on his \rb{bed}.\\
                \bottomrule
            \end{tabularx}
            
        \label{tab:LGCcases}
        \end{table}

\section{The Human Evaluation Setups}
\begin{figure}
        \centering
        \includegraphics{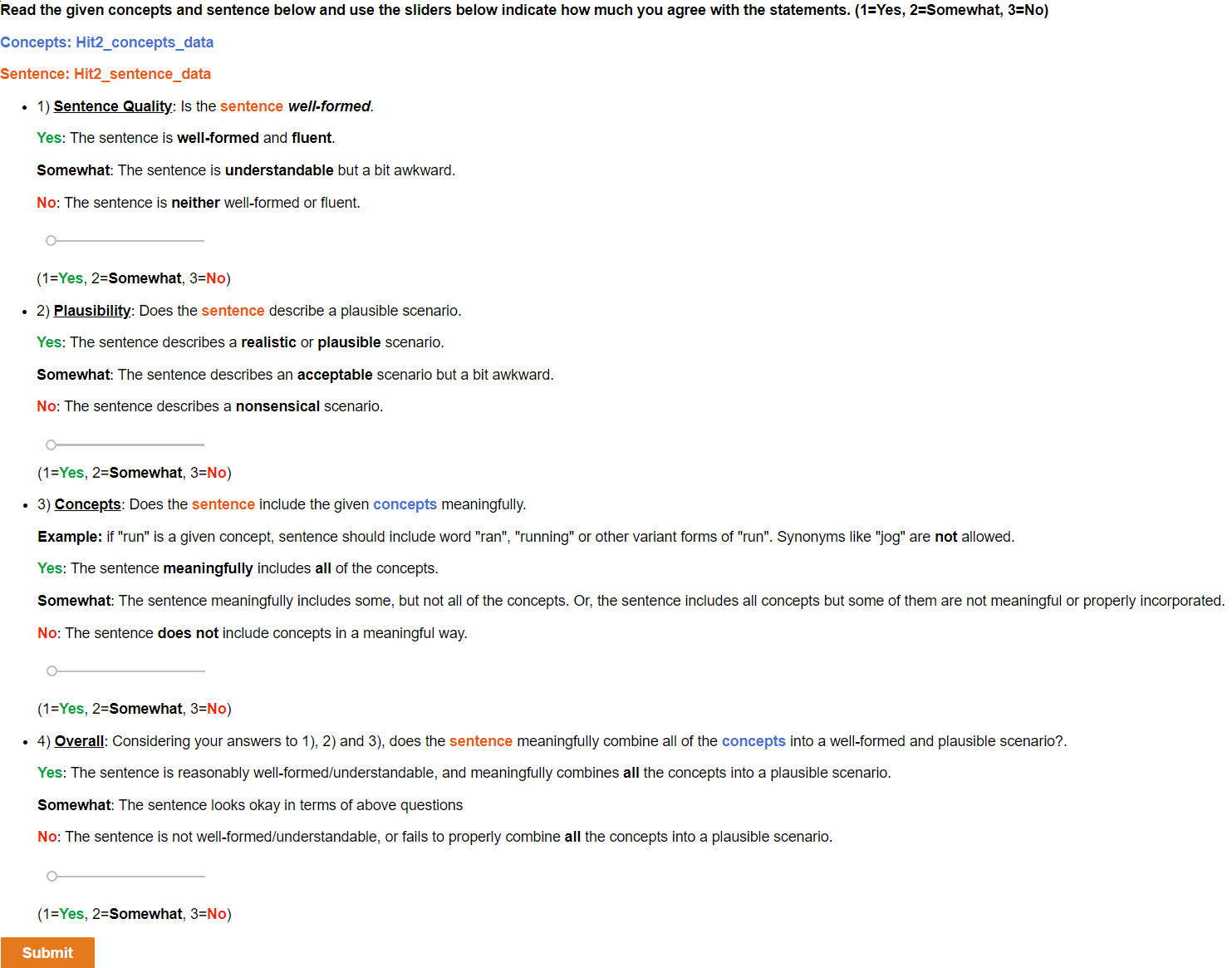}
        \caption{Human evaluation template for LCG experiments. To make the evaluation results easy to read, we score "Yes" for 3 points, "Somewhat" for 2 points and "No" for 1 point. }
        \label{app:fig:template_lcg}
    \end{figure}

    We use the template shown in Fig. \ref{app:fig:template_lcg} to do the human evaluation. We sampled 993 keys from CommonGen test set and generate sentences. For each sentence we ask two MTurkers to annotate. We filter the MTurkers by approval rate greater than $97\%$ and the number of approval greater than $50.$ The pay rate is $\$0.05$ per sentence.
